\documentclass[10pt,twocolumn,letterpaper]{article}

\usepackage[final]{cvpr}

\usepackage{graphicx}
\usepackage[export]{adjustbox}
\usepackage{tabularx}
\usepackage{subcaption}
\usepackage{multirow}
\usepackage{amsmath}
\usepackage{amssymb}
\usepackage{booktabs}
\usepackage{tikz}
\usetikzlibrary{positioning, arrows.meta, shapes.geometric, calc}

\usepackage[pagebackref,breaklinks,colorlinks]{hyperref}

\usepackage[capitalize]{cleveref}
\crefname{section}{Sec.}{Secs.}
\Crefname{section}{Section}{Sections}
\Crefname{table}{Table}{Tables}
\crefname{table}{Tab.}{Tabs.}

\def\cvprPaperID{*****} 
\def\confName{CVPR}
\def\confYear{2023}

\begin{document}

\title{TriLoRA: Integrating SVD for Advanced Style Personalization in Text-to-Image Generation}


\author{
    Chengcheng Feng\textsuperscript{1}\thanks{These authors contributed equally to this work and should be considered co-first authors.}\quad  
    Mu He\textsuperscript{1}\footnotemark[1] \quad 
    Qiuyu Tian\textsuperscript{2} \quad 
    Haojie Yin\textsuperscript{3} \quad 
    Xiaofang Zhao\textsuperscript{1} \thanks{Corresponding author: zhaoxf@ict.ac.cn} \quad 
    Hongwei Tang\textsuperscript{1} \\\quad 
    Xingqiang Wei\textsuperscript{4} \\
    \textsuperscript{1}Institute of Computing Technology, Chinese Academy of Sciences \\
    \textsuperscript{2}Southeast University \\
    \textsuperscript{3}Institute of Automation, Chinese Academy of Sciences \\
    \textsuperscript{4}Shaanxi Normal University \\
}





\maketitle

\begin{abstract}
As deep learning technology continues to advance, image generation models, especially models like Stable Diffusion, are finding increasingly widespread application in visual arts creation. However, these models often face challenges such as overfitting, lack of stability in generated results, and difficulties in accurately capturing the features desired by creators during the fine-tuning process. In response to these challenges, we propose an innovative method that integrates Singular Value Decomposition (SVD) into the Low-Rank Adaptation (LoRA) parameter update strategy, aimed at enhancing the fine-tuning efficiency and output quality of image generation models. By incorporating SVD within the LoRA framework, our method not only effectively reduces the risk of overfitting but also enhances the stability of model outputs, and captures subtle, creator-desired feature adjustments more accurately. We evaluated our method on multiple datasets, and the results show that, compared to traditional fine-tuning methods, our approach significantly improves the model's generalization ability and creative flexibility while maintaining the quality of generation. Moreover, this method maintains LoRA's excellent performance under resource-constrained conditions, allowing for significant improvements in image generation quality without sacrificing the original efficiency and resource advantages.
\end{abstract}

\section{Introduction}

Diffusion models \cite{ho2020denoising,rombach2022high,saharia2022photorealistic}have achieved impressive image generation quality through their outstanding understanding of diverse artistic concepts and enhanced controllability due to support for multimodal conditions, with text being the most popular mode. The usability and flexibility of generative models have further increased with the development of various personalization approaches, such as DreamBooth\cite{ruiz2023dreambooth} and StyleDrop\cite{sohn2023styledrop}. These methods fine-tune the underlying diffusion model on images of specific concepts to produce novel representations in various contexts. These concepts can be specific objects or people, or an artistic style.

While personalized methods have been deployed to independently tackle themes and styles, the persistent challenges of overfitting and training stability require urgent solutions. Traditional techniques, though somewhat effective in mitigating these issues, often fall short during extremely brief or prolonged training sessions, thus constraining the model's efficacy and applicability. For instance, users might seek to incorporate style elements from other designs into an existing garment style, aiming to do so while maintaining the original style to a reasonable extent. Addressing this issue could significantly broaden the creative freedom afforded to creators, offering a more extensive space for parameter tuning. This, in turn, would allow for a more diverse array of effective generation outcomes, thereby providing users with more options. Envisioning the large-scale deployment of such a tool—comprising a library rich in styles and themes—presents an open research question. The goal is to enable the task of rendering various subjects in any style at will, pushing the boundaries of current capabilities and exploring new frontiers in generative models' applicability and versatility.

Parameter-Efficient Fine-Tuning (PEFT) methods\cite{hu2021lora,zhang2022adaptive,li2021prefix,liu2021p,liu2023gpt,lester2021power,liu2022few,sanh2021multitask,hyeon2021fedpara,edalati2022krona,li2023loftq,qiu2024controlling,sung2021training} allow for concept-driven personalized fine-tuning of models with a lower memory and storage budget. Among various PEFT methods, Low-Rank Adaptation (LoRA)\cite{hu2021lora} has emerged as the preferred approach for researchers and practitioners due to its versatility. LoRA learns a low-rank decomposition of the weight matrices for the attention layers (these learned weights are often referred to as "LoRAs"). By combining LoRA with algorithms such as DreamBooth\cite{ruiz2023dreambooth}, the LoRA weights learned for specific subjects enable the model to generate themes with semantic variations, allowing for the creation of detailed and semantically rich outputs.

As LoRA personalization becomes increasingly popular, attempts are being made to incorporate various style elements into "LoRAs." To achieve the desired effect, users often try to control the "strength" of each LoRA by adjusting parameters. Sometimes, it requires careful grid searching and subjective human assessment to find a parameter combination that allows for the incorporation of specific style elements into existing themes for a more distinct style change. However, using LoRA makes it difficult to control the degree of integration of feature elements, leading to issues where style features may not be pronounced enough or features may be too prominent.

In our work, we introduce TriLoRA that offers computational and memory costs similar to LoRA-type methods, but significantly improves accuracy under the same parameter and computational budget, while also being easier to use and adjust. Specifically, in practical experiments, it has proven more stable than traditional LoRA and can reduce the likelihood of image overfitting in extremely high iteration training, greatly increasing the freedom for creators during LoRA training. Additionally, it achieves better results in training with fewer iterations compared to LoRA. Overall, our method not only increases the flexibility of LoRA but also enhances its stability.

Our approach leverages the easily modifiable LoRA training framework provided by the diffusers library\cite{von-platen-etal-2022-diffusers} and incorporates the concept of singular value decomposition (SVD)\cite{bisgard2020analysis} into the original idea of low-rank decomposition. SVD is a decomposition technique that breaks down any matrix into the product of three specific matrices: an orthogonal matrix, a diagonal matrix (whose diagonal elements are singular values), and the transpose of another orthogonal matrix. Specifically, we propose a scheme that trains two adapters: a standard low-rank adapter (LoRA) and an additional, smaller adapter, both of which are trained "in parallel" relative to the original pretrained weights. There are three challenges we must address: 1) identifying a high-performance small adapter, 2) finding a co-training mechanism that yields stable convergence, and 3) determining a set of parameters that are suitable for most use cases.

Based on previous work in the field, we have addressed all three challenges and demonstrated that the TriLoRA adapter can significantly enhance the robustness of the resultant models under comparable parameter, memory, and computational budgets relative to standard adapters. We complement our algorithmic contribution with an efficient system implementation of TriLoRA in Pytorch, which runs swiftly on NVIDIA GPUs.

In summary, we provide promising evidence that our method significantly improves robustness over LoRA in most diffusion model use scenarios, enhancing practical usability. Therefore, TriLoRA can become an additional technique in the creator's toolkit for using diffusion models in resource-constrained environments.

\section{Related Work}
\subsection{Parameter-Efficient Fine-Tuning.} 

Recent advancements in Large Language Models (LLMs) \cite{touvron2023llama,zhang2022opt} have showcased exceptional capabilities across a spectrum of NLP tasks. However, the significant memory and computational demands associated with these models present substantial challenges in both training and inference phases. Despite efforts to mitigate computational demands, memory constraints persist as a notable bottleneck\cite{min2021metaicl,wei2021finetuned,ouyang2022training,wang2022super,liu2022few}. 

In response to the significant computational challenges presented by large-scale pre-trained models such as GPT\cite{openaiIntroducingChatGPT}, BERT\cite{devlin2018bert}, and Vision Transformers\cite{dosovitskiy2020image}, Parameter-Efficient Fine-Tuning (PEFT) has emerged as a key strategy. Distinguished from traditional full fine-tuning, PEFT prioritizes targeted fine-tuning on smaller downstream tasks rather than initiating full-scale training from scratch, thereby substantially reducing computational costs. This method not only diminishes the demand for extensive computational resources and time but also establishes PEFT as a more feasible solution across a broad range of research and application scenarios\cite{hu2021lora,zhang2022adaptive,li2021prefix,liu2021p,liu2023gpt,lester2021power,liu2022few,sanh2021multitask,hyeon2021fedpara,edalati2022krona,li2023loftq,qiu2024controlling,sung2021training}.Innovations in PEFT, including Adapter Layers, Prompt Tuning, BitFit, and Low-Rank Adaptation (LoRA), focus on modifying a minimal subset of model parameters, thus facilitating task adaptation while preserving the original model architecture and knowledge base.

Notably, within the realm of image generation models, such as the Stable Diffusion models\cite{rombach2022high}, LoRA\cite{hu2021lora} has been recognized for its efficacy in tailoring models to generate images with specific styles or themes through minimal parameter adjustments. Introduced by Hu et al. (2021), LoRA's approach of integrating trainable low-rank matrices with existing layer weights modifies the output of generative models with only a slight increase in trainable parameters. This technique strikes a balance between fostering user creativity and maintaining computational efficiency, proving especially suitable for customized large-scale image creation tasks.

\subsection{Sparse Fine-Tuning.} 

In the realm of large language models, sparse training and fine-tuning have risen to prominence as strategies to mitigate the substantial computational and memory demands these models entail \cite{hoefler2021sparsity}. These approaches are applicable across both the inference \cite{gale2019state,singh2020woodfisher,sanh2020movement,frantar2022optimal} and training phases \cite{evci2020rigging,peste2021ac,hubara2021accelerated,jiang2022exposing,nikdan2023sparseprop}, encompassing the entire model lifecycle. Sparse fine-tuning entails the pre-training and subsequent sparsification of networks on extensive datasets, followed by the application of fine-tuning to tailor the network for precise downstream tasks, all while preserving the established sparsity patterns \cite{nikdan2023sparseprop}.

While both sparse fine-tuning and sparse adaptation strategies focus on refining a select subset of a model's parameters, there is a distinct difference in how they treat the uninvolved parameters. Specifically, sparse fine-tuning disregards the non-selected parameters altogether, effectively removing them from the optimization process. In contrast, sparse adaptation temporarily freezes these parameters during the fine-tuning phase, ensuring they remain unchanged yet still part of the model's structure. This nuance proves especially vital in the domain of image diffusion models, where the quest for nuanced stylistic transformations is common. By selectively fine-tuning individual features, sparse fine-tuning can adeptly meet the demand for specific stylistic modifications. This method offers a focused mechanism for altering images to achieve the desired stylistic outcomes, thus efficiently balancing the use of computational and memory resources in the process.

\subsection{Singular Value Decomposition (SVD)}
Singular Value Decomposition (SVD) represents an indispensable mathematical framework widely applied in the realm of data compression, delivering a systematic strategy to diminish the storage and transmission demands of voluminous datasets while safeguarding vital information. Fundamentally, SVD decomposes a given matrix $A$ of dimensions $m \times n$ into three distinct matrices:
\begin{equation}
A = U_r \Sigma_r V^T
\end{equation}

where:
\begin{itemize}
    \item $U$ is an $m \times m$ orthogonal matrix whose columns are the left singular vectors of $A$.
    \item $\Sigma$ is an $m \times n$ diagonal matrix with non-negative real numbers on the diagonal, known as singular values, sorted in descending order. These singular values represent the intrinsic data dimensions, capturing the essence of the data's variance and importance.
    \item $V_T$ is the transpose of an $n \times n$ orthogonal matrix whose columns are the right singular vectors of $A$.
\end{itemize}

The decomposition highlights the fundamental structure of the data, where the singular values in $\Sigma$ quantify the contribution of each singular vector towards the data's composition. This characteristic makes SVD uniquely suited for data compression—by retaining only the top $k$ largest singular values (and their corresponding singular vectors in $U and V_T$), one can construct a lower-rank approximation $\hat{A} = U_k \Sigma_k V_k^T$ of the original matrix. This approximation significantly reduces the amount of data needed to represent $A$ thus achieving compression.

SVD's prowess in data compression emanates from its proficiency in discerning and preserving the data's most pivotal features, rendering it especially beneficial for uses in image and signal processing. Through the exclusion of singular values (and vectors) contributing minimally to the dataset’s structure, SVD facilitates a form of lossy compression whereby the decompressed data retains the core characteristics of the original dataset with negligible quality degradation perceptible.

For example, within the context of image compression, SVD trims the dimensions of image files by sustaining only a select array of singular values that encapsulate the bulk of the visual data within the image. This mechanism adeptly captures the image’s significant features while optimally minimizing superfluous parameters, thus preserving an image rendition visually akin to the original.

Moreover, the utility of SVD in data compression transcends image data to encompass text, audio, and video data, underscoring its adaptability and efficacy across a diverse spectrum of media types. The mathematical sophistication and computational practicability of SVD have solidified its status as a foundational technique in data compression, striking a critical balance between compression efficiency and quality preservation vital for effective data management and transmission in today's digital era.

\section{Method}
\subsection{Low-Rank Adaptation}
Stable Diffusion is a type of pretrained model that has been trained on a large amount of data, hence the original model often possesses good generalization and practicality. LoRA fine-tuning targets the model's linear transformation layers, such as the self-attention layers and feedforward neural network layers in Transformer models. The key is to introduce low-rank matrices to modify the model's weights instead of directly training the entire weight matrix. The general principle is as follows:

Let the original weight matrix of the linear transformation layer be $W_0 \in \mathbb{R}^{d \times k}$, where $d$ and $k$ represent the dimensions of the input and output, respectively. LoRA fine-tuning modifies $W$  by adding a parameter update matrix $\delta W$, which is decomposed into two low-rank matrices $A \in \mathbb{R}^{d \times r}$ and $B \in \mathbb{R}^{r \times k}$  for efficient parameter updating. Here, $r$ is a rank much smaller than $d$ and $k$, typically $r \ll min(d,k)$ Therefore, the linear transformation after LoRA fine-tuning is represented as:
\begin{align}
    W' = W_0 + \Delta W \\
    W' = W_0 + AB
\end{align}
where:
\begin{itemize}
    \item $W$ represent the original weight matrix 
    \item $AB$ represents the low-rank update.
\end{itemize}

\subsection{Compact Singular Value Decomposition (Compact
SVD)}
Compact Singular Value Decomposition (Compact
SVD) is a powerful variant of the classical SVD, tailored
for more efficient computation and storage, especially in the
case of sparse or low-rank matrices. Unlike standard SVD,
Compact SVD focuses on reducing the dimensions of these
matrices based on the rank r of the original matrix.
The essence of Compact SVD lies in its ability to ef-
ficiently approximate the original matrix by leveraging its
inherent rank. Similarly, the compact SVD of a matrix $W$ is
given by:
\begin{equation}
    W = U_r \Sigma_r V^T
\end{equation}
where:
\begin{itemize}
    \item $r\ll min(m, n)$
    \item $U_r$ is simplified orthogonal matrices of sizes $d\times r$,
    \item $V_r$ is simplified orthogonal matrices of sizes $k\times r$,
    \item $\Sigma_r$ s an $r \times r$ diagonal matrix, containing only the r
largest singular values.
\end{itemize}

This reduction not only captures the most significant fea-
tures of the original matrix but also significantly reduces
computational complexity and storage requirements.
The applications of Compact SVD span multiple fields,
from image processing and signal compression to machine
learning and data mining. In these domains, the ability to
efficiently approximate and process large datasets or matri-
ces is crucial. For example, in image compression, Compact
SVD retains only the most critical parts of the image, sig-
nificantly reducing the amount of data, facilitating storage
and transmission, while not severely affecting image qual-
ity. This characteristic is greatly beneficial in fine-tuning
the checkpoints in large parameter diffusion model. By fo-
cusing on the parts of the dataset with the highest amount of
information, Compact SVD can simplify data, maintaining
its essence while enhancing the manageability of computa-
tions.

\begin{figure}[tp] 
  \centering 
  \begin{tikzpicture}[
  node distance = 1cm and 0.5cm,
  box/.style = {draw, thick, align=center, fill=blue!30, minimum height=1cm, text width=3.5cm},
  widebox/.style = {draw, thick, align=center, fill=gray!30, text width=4.5cm, minimum height=1cm},
  multiply_node/.style = {draw, thick, inner sep=0pt, circle, fill=white},
  arrow_style/.style = {thick, -Stealth},
  trapezium_node1/.style = {draw, thick, trapezium, trapezium stretches body, fill=orange!30, minimum width=3cm, trapezium left angle=70, trapezium right angle=70, align=center},
  trapezium_node2/.style = {draw, thick, trapezium, trapezium stretches body, fill=orange!30, minimum width=3cm, trapezium left angle=110, trapezium right angle=110, align=center},
  rectangle_node/.style = {draw, thick, align=center, fill=orange!30, minimum width=2.5cm, minimum height=1cm} ,
  ]

    \node[widebox] (w0) {Frozen Pretrained Weight\\ $W_0 \in \mathbb{R}^{p \times q}$};
    \node[box, below=3cm of w0] (h) {$h$};
    \node[box, above=3cm of w0] (h-prime) {$h'$};

    \node[trapezium_node1] (a) at ($(w0)!0.4!(h) + (4.5cm, 0)$) {$A \in \mathbb{R}^{r_1 \times q}$};
    \node[rectangle_node] (c) at ($(w0)!0.05!(h-prime) + (4.5cm, 0)$) {$C \in \mathbb{R}^{r_2 \times r_1}$};
    \node[trapezium_node2] (b) at ($(c)!0.5!(h-prime) + (2.3cm, 0)$) {$B \in \mathbb{R}^{p \times r_2}$};
    \node[multiply_node] (mult1) at ($(a)!0.5!(c)$) {$\times$}; 
    \node[multiply_node] (mult2) at ($(c)!0.5!(b)$) {$\times$}; 

    \draw[arrow_style] (h) -- (w0);
    \draw[arrow_style] (w0) -- (h-prime);
    \draw[arrow_style] (h) -- (a);
    \draw[arrow_style] (b) -- (h-prime);

    \draw[thick] (a) -- (mult1);
    \draw[thick] (mult1) -- (c);
    \draw[thick] (c) -- (mult2);
    \draw[thick] (mult2) -- (b);

\end{tikzpicture}
  \begin{tikzpicture}
  \end{tikzpicture}
  \caption{Structural Design of the TriLoRA Model.} 
  \label{TriLoRA} 
\end{figure}
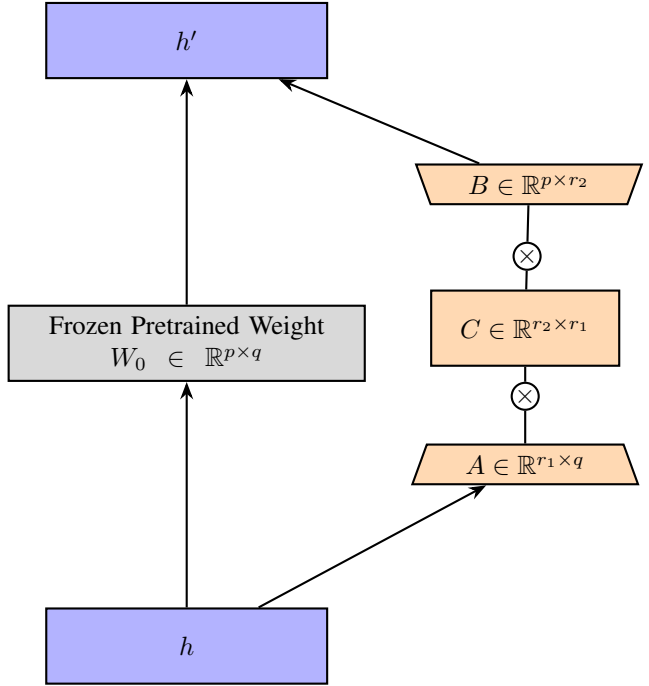

\subsection{TriLoRA}
In past usage, it's been noted that the choice of rank in LoRA often appears arbitrary. Based on research by Hu E. J. and colleagues, it's found that in LoRA's low-rank decomposition, often the first 64 features have a significant impact on the parameter matrix, indicating that $r \leq 64$ can effectively fine-tune the model. However, in the process of generating images with diffusion models, creators often have very specific features they wish to alter, intending to integrate desired elements and thus modify the overall image style. Based on these findings, We introduce TriLoRA, a novel extension of the Low-Rank Adaptation (LoRA) technique, tailored for enhancing the self-attention mechanism within neural network architectures, particularly those utilized in generative tasks such as text-to-image generation. Compact SVD determines the number of features $r$ that the creator is focused on, thus providing a relatively accurate selection space.

In the TriLoRA framework, the attention mechanism is modified by integrating Compact SVD into LoRa shown in \cref{TriLoRA}, further optimizing the update of the weight matrix. Unlike traditional LoRa, where the update $W_0 + \Delta W$ is represented by $W_0 + AB$, we introduce the idea of Compact SVD, representing $\Delta W$  as the product of three matrices: 
\begin{align}
    \Delta W = U \Sigma V^T
\end{align}
i.e.
\begin{align}
    W_{trilora} = W_0+ U \Sigma V^T
\end{align}
where:
\begin{itemize}
    \item $U$ and $V$ are unitary orthogonal matrices;
    \item $\Sigma$ is a diagonal matrix containing the singular values.
\end{itemize}
This step allows us to capture the features in the weight update process more finely, thereby improving the model's adaptability while reducing the number of parameters required.

In implementation, $U$ is initialized using Gaussian initialization, $\Sigma$ with SVD initialization to retain the desired singular values, and $V^T$ with zero initialization. This choice of initialization strategy is based on providing a balanced starting point that retains key information from the pretrained model while offering sufficient flexibility for learning during the model adaptation process. By appropriately initializing and subsequently optimizing these matrices, we achieve maximal adaptability to new tasks while maintaining relatively low model complexity.

Similarly, during training, $W_0$ is fixed and does not participate in updates, while $U, \Sigma$ and $V^T$ serve as trainable parameters in the model's optimization. This approach allows us to finely adjust a small number of parameters rather than retraining the entire pretrained model when adapting to new tasks, significantly reducing the computational costs and time expenses of model adaptation.

Moreover, to simplify calculations, we do not strictly follow the SVD process but merely obtain the $\Delta W$ matrix containing the singular values. while retaining only $r$ desired singular values.

TriLoRA's methodology presents a refined approach to enhancing the self-attention mechanism's capacity for nuanced information processing. By leveraging triple directional low-rank matrices, it allows for a tailored modulation of attention dynamics, promising improvements in the model's generative capabilities. This technique not only extends the flexibility and efficiency of the attention mechanism but also opens avenues for further exploration into advanced neural network optimization strategies.

\section{Experiments}
\subsection{Datasets Preparation}

This section focuses on the datasets prepared for evaluating the adaptability of TriLoRA and LoRA within the Stable Diffusion model to specific styles or themes.

\noindent \textbf{Dataset Construction.}
We constructed two datasets: one focusing on an array of fantastical creatures and the other on a specific style of clothing. The first dataset, known as the Pokemon\cite{pinkney2022pokemon} dataset,  featuring a diverse collection of 736 Pokemon images. This dataset allows us to train models to recognize and generate a wide variety of creature designs, each with distinct attributes and colors. The second dataset, named the GAC dataset, was collected from the Giorgio Armani website \footnote{\href{https://www.armani.com/en-us/giorgio-armani/woman/clothing}{https://www.armani.com/en-us/giorgio-armani/woman/clothing} (Accessed: 2024-01-18)}, focusing on the dress category of clothing flat images, totaling 52 pictures. 

\noindent \textbf{Label Generation.}
For the clothing-focused GAC dataset, preliminary annotations were conducted using GPT's API and few-shot learning techniques, which were then meticulously verified manually to guarantee label accuracy and reliability.  Emphasis was placed on clothing attributes to enrich the model's understanding of fashion-related elements.In contrast, the Pokemon dataset came with its own set of predefined labels, thereby eliminating the need for additional GPT-based annotation. By using unconventional letter combinations as trigger words, our goal was to analyze the content learned by the model, ensuring it can differentiate between the knowledge acquired from pre-trained model and that obtained through training with LoRA and TriLoRA.

\subsection{Experimental Setting}

\begin{figure*}[ht!]
\setkeys{Gin}{width=\linewidth,keepaspectratio}
\centering
\captionsetup{justification=raggedright}
\begin{tabularx}{\textwidth}{c*{3}{X}}
  & \centering blue and white, ankle-length, cloud print, strapless, draped bodice, asymmetric skirt with drape side detail, lined
  & \centering grey, knee-length, solid color, shawl collar, sleeveless, wrap design, belt closure at waist, unlined
  & \centering black and silver, knee-length, sequin bodice, fringe skirt, sleeveless, tuxedo lapel, no closure visible, lined \tabularnewline
  \multirow{20}{*}[1.5in]{LoRA\cite{hu2021lora}}
  & \includegraphics[valign=m]{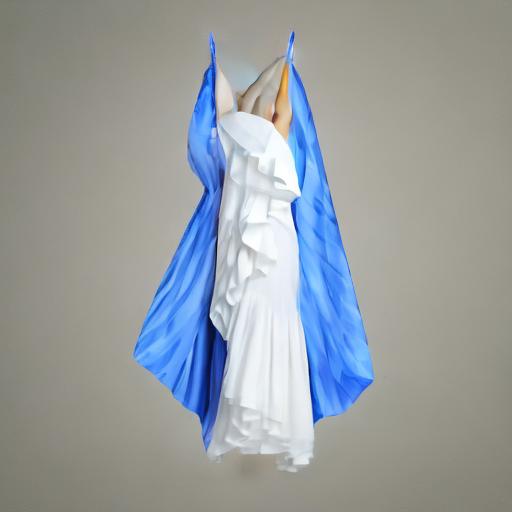}
  & \includegraphics[valign=m]{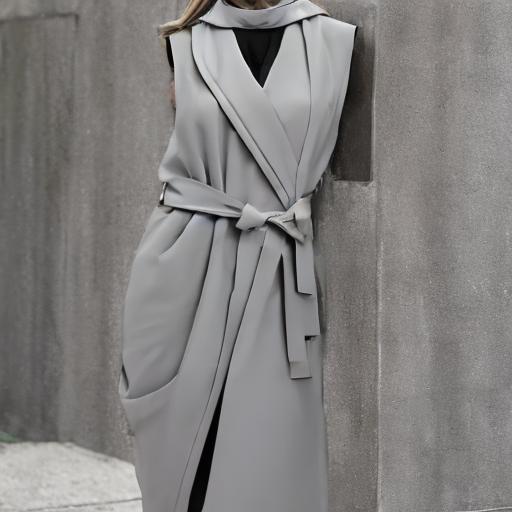}
  & \includegraphics[valign=m]{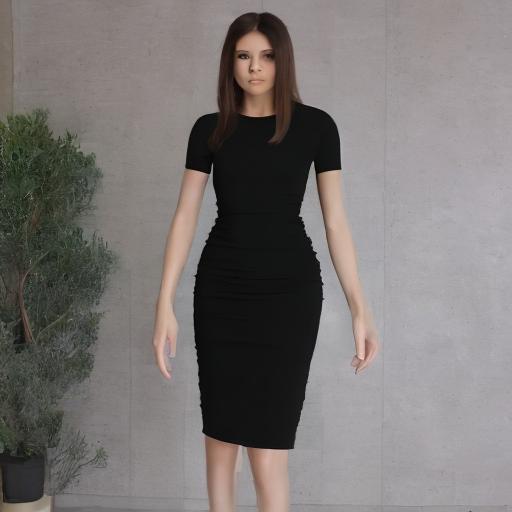} \tabularnewline
  \multirow{20}{*}[1.5in]{TriLoRA(Ours)}
  & \includegraphics[valign=m]{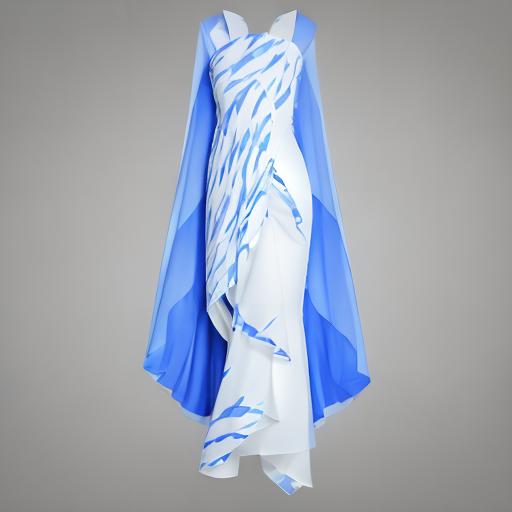}
  & \includegraphics[valign=m]{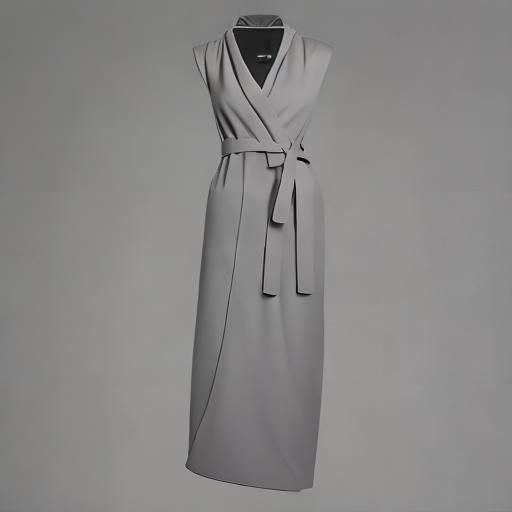}
  & \includegraphics[valign=m]{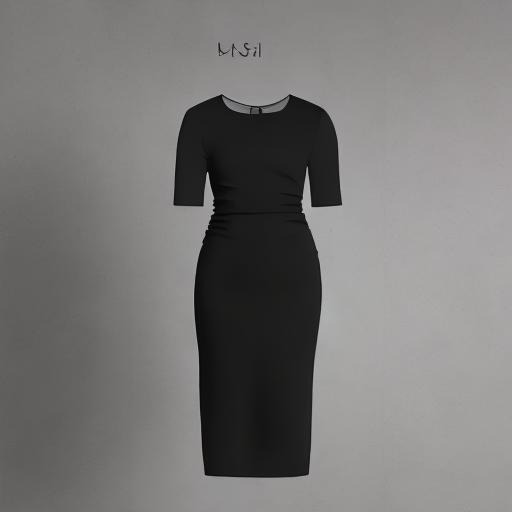} \tabularnewline
\end{tabularx}
\caption{Images generated by LoRA and TriLoRA after training for 10 epochs on the GAC dataset, with responses to three prompts (each column).}
\label{armani_effect}
\end{figure*}

\noindent \textbf{Evaluation Metrics.}
To comprehensively assess the TriLoRA model's performance, we adopted two principal quantitative metrics: the Normalized Fréchet Inception Distance\cite{heusel2017gans} (Normalized FID,NFID) and the CLIP Score\cite{radford2021learning}.

\begin{itemize}
	\item \textbf{Normalized FID Score} is employed to measure the visual quality of generated images by comparing them to real images. By normalizing the standard FID Score, Normalized FID Score adjusts for differences in dataset size and composition, ensuring a fair comparison across various settings. 
	\item \textbf{CLIP Score} assesses the semantic congruency between the textual descriptions and the generated images, showcasing the model's ability to accurately interpret and visualize textual prompts.
\end{itemize}

Furthermore, a \textbf{user study} was conducted to supplement the quantitative metrics with qualitative insights. Through visual evaluation, participants provided feedback on the realism and fidelity of the generated images. 

\noindent \textbf{Baselines.}
In our experiments involving the Pokemon and GAC datasets, the pre-trained Stable Diffusion 1.5\cite{huggingfaceRunwaymlstablediffusionv15Hugging}(SD V1.5) model was selected as the benchmark. The Normalized FID scores were computed against this baseline to objectively measure enhancements in image synthesis quality attributed to our proposed methods. To ensure consistency in our results, the inference setup was fixed with a seed=30, num\_inference\_steps=20 to afford detailed image progression, and guidance\_scale=7.5 to calibrate the textual prompt's influence on image generation.

Contrasting with LoRA or TriLoRA, the pre-trained model's trigger words corresponded to specific real-world vocabulary intended to direct image generation. For instance, within the Pokemon dataset, the trigger phrase "Pokemon" was employed, whereas for the GAC dataset, "Armani" served as the prompt. These prompts guided the pre-trained model to generate images closely associated with the intended subjects, facilitating a targeted generation process for comparative evaluation.

\noindent \textbf{Implemention.}
The implementation of TriLoRA was based on the diffusers\cite{von-platen-etal-2022-diffusers} library, version 0.24.0. All experiments were conducted on a single Nvidia V100 GPU.  A fixed learning rate was applied across all experiments for standardization and comparability of results.

Specifically for the GAC dataset, we set the learning rate to $5*10^{-5}$ .  We fine-tuned the Hugging Face portrait model, Realistic Vision\cite{huggingfaceSG161222Realistic_Vision_V51_noVAEHugging}, using over 1000 clothing flat images through DreamBooth\cite{ruiz2023dreambooth}, which yielded a specialized model for clothing flat images.  Subsequently, TriLoRA and LoRA were each fine-tuned on top of this tailored clothing flat image model, spanning 10 epochs.

For the Pokemon dataset, the learning rate was  set to $10^{-4}$.   Training was conducted using the anime-based pre-trained model  AnyLoRA\cite{huggingfaceLykonAnyLoRAHugging} as a foundation, spanning a total of 100 epochs.   Each training iteration involved the random selection of 100 images from the dataset, facilitated by the max\_train\_samples parameter to ensure diverse exposure to the model's learning process.

\subsection{Qualitative Results}

\begin{figure*}[ht!]
\setkeys{Gin}{width=\linewidth,keepaspectratio}
\centering
\captionsetup{justification=raggedright}
\begin{tabularx}{\textwidth}{c*{3}{X}}
  & \centering a cartoon bird with multicolored wings and a black beak
  & \centering a drawing of a pokemon pikachu in a bowl
  & \centering a drawing of a white and blue pokemon \tabularnewline
  \multirow{20}{*}[1.5in]{LoRA\cite{hu2021lora}}
  & \includegraphics[valign=m]{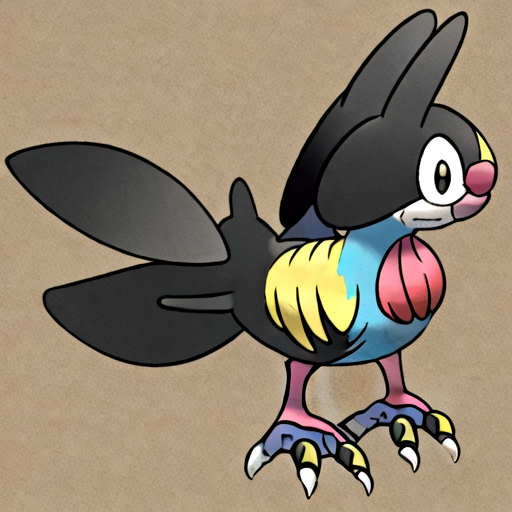}
  & \includegraphics[valign=m]{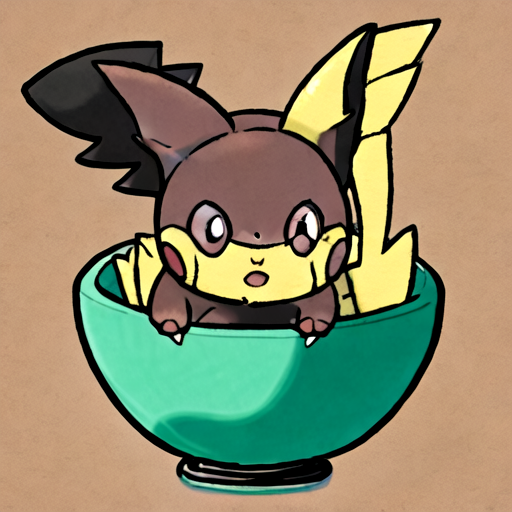}
  & \includegraphics[valign=m]{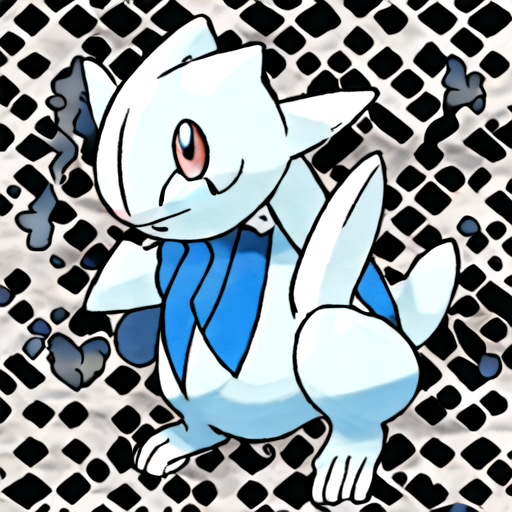} \tabularnewline
  \multirow{20}{*}[1.5in]{TriLoRA(Ours)}
  & \includegraphics[valign=m]{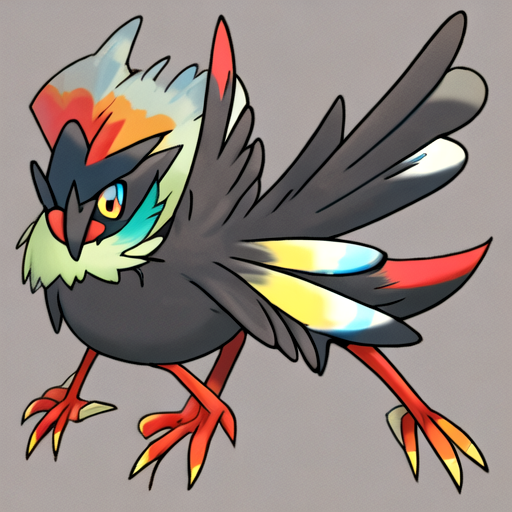}
  & \includegraphics[valign=m]{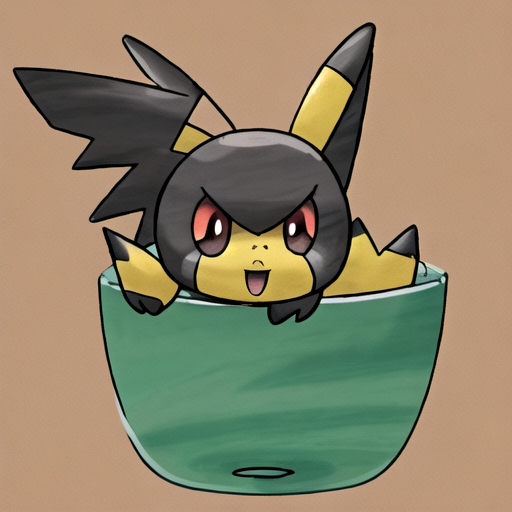}
  & \includegraphics[valign=m]{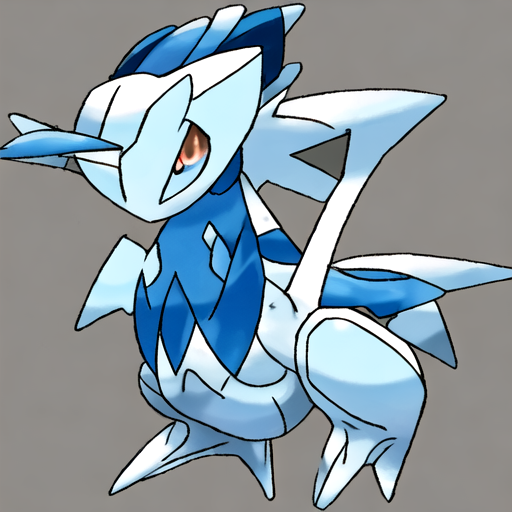} \tabularnewline
\end{tabularx}
\caption{Images generated by the LoRA and TriLoRA models on the Pokemon\cite{pinkney2022pokemon} dataset after 100 training epochs, corresponding to distinct prompts across columns.}
\label{pokemon_effect}
\end{figure*}

\cref{armani_effect} displays the generated images by LoRA and TriLoRA models after fine-tuning for 10 epochs on the GAC dataset  . It is observed that the images produced by TriLoRA generally possess a superior visual aesthetic when compared to those generated by LoRA. With the first prompt, LoRA's rendition partly breaks down in detail and fails to establish a clear correlation between the cloud print and the depicted garment, a distinction that TriLoRA manages to capture effectively. Additionally, for the second and third prompts, LoRA not only incorrectly includes partial or entire person, which contradicts the flat-lay garment concept, but it also struggles to replicate the solid-colored background, compromising the visual cleanliness expected in flat-lay photography. Conversely, TriLoRA adapts to produce imagery adhering to the flat-lay format and accurately renders the solid-colored background, upholding the integrity of clothing presentation.

\cref{pokemon_effect} presents a comparative visualization of images generated by TriLoRA and LoRA after 100 epochs of training with identical parameters on the Pokemon dataset, each model responding to the same set of prompts. The comparative study reveals that TriLoRA exhibits a more sophisticated artistic style and produces images with greater visual stability. For instance, in the imagery corresponding to the first prompt, "a cartoon bird with multicolored wings and a black beak", LoRA fails to accurately manifest the prompt's description—yielding an image with black wings and a misidentified red beak-like element. In stark contrast, TriLoRA adeptly generates an image that closely aligns with the given prompt. Furthermore, when assessing the images for the second and third prompts, LoRA's results are noticeably less refined in terms of artistic complexity compared to TriLoRA. Notably, for the third prompt, LoRA struggles to replicate the pure-colored background that is a staple of the dataset, in contrast to TriLoRA"s consistent performance in this regard.

\subsection{Quantitative Results}

\begin{table}[htbp]
\centering
\caption{Quantitative Evaluation of Image Synthesis on GAC and Pokemon Datasets. CLIP score\cite{radford2021learning} was utilized for assessing text-image consistency, while the Normalized FID\cite{heusel2017gans}  was calculated with Stable Diffusion 1.5 serving as the baseline to evaluate the visual fidelity of generated images.}
\begin{tabular}{@{}lcccc@{}}
\toprule
& \multicolumn{2}{c}{GAC} & \multicolumn{2}{c}{Pokemon\cite{pinkney2022pokemon}}  \\ 
\cmidrule(lr){2-3} \cmidrule(lr){4-5}  
& NFID ↓ & CLIP ↑ & NFID ↓ & CLIP ↑ \\ 
\midrule
SD V1.5\cite{huggingfaceRunwaymlstablediffusionv15Hugging} & 1 & 32.2 & 1 & 29.2   \\
LoRA\cite{hu2021lora} & 0.923 & \textbf{32.5} & 0.867 & \textbf{30.3}  \\
TriLoRA(Ours) & \textbf{0.833} & \textbf{32.5} & \textbf{0.796} & 30.1  \\
\bottomrule
\end{tabular}
\label{comparison}
\end{table}

For the GAC dataset, inference was conducted using 52 distinct prompts describing dress attributes. Given the discrepancy in aspect ratios between the original clothing images and the generated images, both sets were normalized to a resolution of 512x512 pixels to satisfy the resolution consistency requirement for FID computation. For the Pokemon dataset, image generation was performed using all 736 prompts describing the creatures depicted in the dataset. \cref{comparison} delineates the quantitative outcomes obtained after training TriLoRA and LoRA for 10 epochs on the GAC dataset and 100 epochs on the Pokemon dataset, respectively. The Normalized FID calculations were benchmarked against Stable Diffusion 1.5. The results indicate a lower Normalized FID for TriLoRA compared to LoRA, signifying superior performance. Moreover, the CLIP Scores across Stable Diffusion 1.5, LoRA, and TriLoRA exhibit minimal variance. This observation could be attributed to the modification targeting the UNet architecture within LoRA, where the text encoder CLIP's parameters were frozen prior to training, with TriLoRA introducing further enhancements upon this foundation.

\begin{table}
  \centering
  \begin{tabular}{@{}lc@{}}
    \toprule
    Model & Pokemon\cite{pinkney2022pokemon}   \\
    \midrule
    SD V1.5\cite{huggingfaceRunwaymlstablediffusionv15Hugging} & 42.1\% \\
    LoRA\cite{hu2021lora} & 63.5\% \\
    TriLoRA(Ours) & \textbf{71.3}\%\\
    \bottomrule
  \end{tabular}
  \caption{User Study on Textual-Visual Consistency and Visual Attractiveness from 100 Pokemon Prompts.}
  \label{user_study}
\end{table}

\noindent \textbf{User Study. }In our user study, 15 participants were asked to evaluate images generated from 50 randomly selected prompts from the Pokemon dataset.  Participants provided ratings that captured both textual-visual consistency and the visual attractiveness of each image. As illustrated in \cref{user_study}, the results of this user study suggest that TriLoRA outperforms its counterparts, achieving the highest scores in both evaluated dimensions.  This indicates a superior ability of TriLoRA to generate visually pleasing images that closely align with the textual prompts, as perceived by the study participants.

\section{Discussions}

The introduction of an additional matrix in TriLoRA, enhancing the LoRA model, significantly improves image stability.   However, this added complexity extends the time required for the model to converge, necessitating more training epochs for optimal performance.

\begin{figure}[h!]
  \centering
  \begin{subfigure}{.25\textwidth}
    \centering
    \includegraphics[width=.9\linewidth]{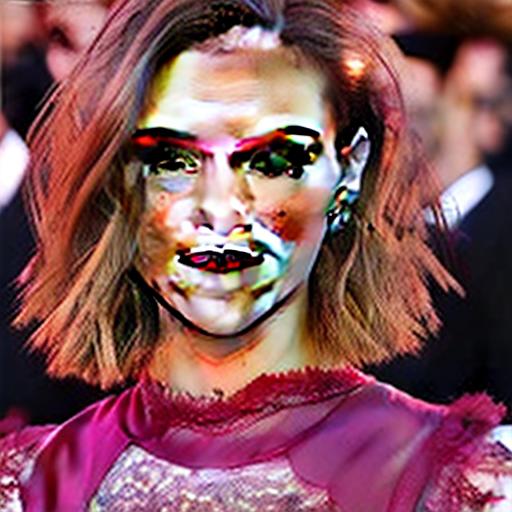}
    \caption{LoRA\cite{hu2021lora}}
  \end{subfigure}%
  \begin{subfigure}{.25\textwidth}
    \centering
    \includegraphics[width=.9\linewidth]{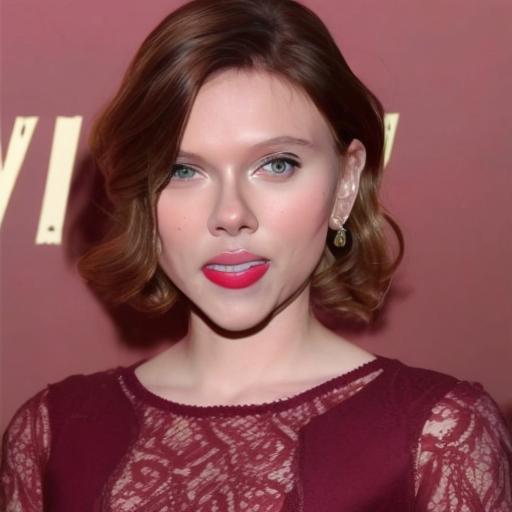}
    \caption{TriLoRA(Ours)}
  \end{subfigure}
  \caption{Comparative Synthesis by LoRA and TriLoRA After 500 Training Epochs on Scarlett Johansson\cite{kaggleCelebrityFace} Dataset with epiCRealism\cite{huggingfaceEmilianJRepiCRealismHugging} Pre-trained Model. Prompt: "Neutral expression, evening makeup, light brown choppy bob hairstyle, stud earrings, burgundy lace dress, event background." Note the significant presence of bright artifacts in the image produced by LoRA.}
  \label{sj}
\end{figure}

An important finding from our comprehensive testing is the reduced incidence of artifacts, such as unwarranted brightness spots indicative of overfitting, in TriLoRA compared to LoRA.   For example, while training on the Scarlett Johansson \cite{kaggleCelebrityFace} dataset for 1000 epochs, LoRA manifested such artifacts by the 500th epoch, whereas TriLoRA remained unaffected, suggesting a more robust optimization capacity and an enhanced resistance to overfitting, as depicted in \cref{sj}.

Despite these advancements, TriLoRA and LoRA share inherent limitations due to their reliance on the quality of pre-trained models.   Our experiments indicate that although TriLoRA achieves FID scores than LoRA, reflecting closer fidelity to the original data distribution, the enhancement in visual quality is not markedly improved when utilizing pre-trained models of only moderate quality.

This underscores the critical balance between model complexity, training dynamics, and pre-trained model quality.   Addressing the dependency on pre-trained models and elevating the perceptual quality of the output images are paramount for the advancement of TriLoRA applicability.   Future efforts should therefore aim at refining model architectures and training methodologies to alleviate these constraints, thereby propelling TriLoRA towards more realistic and high-quality image generation.

\section{Conclusion}
Our study presents a novel integration of Singular Value Decomposition (SVD) with Low-Rank Adaptation (LoRA) to refine the fine-tuning of image generation models like Stable Diffusion. This approach effectively counters challenges like overfitting and output instability, while enhancing the model's ability to capture nuanced, creator-specific features. The empirical results across various datasets underscore our method's superiority in improving model generalization and creative expression, without compromising efficiency or resource constraints. This advancement not only elevates the quality of generated images but also expands the creative canvas for artists, offering them greater flexibility and freedom in their artistic endeavors. 

{\small
\bibliographystyle{ieee_fullname}
\bibliography{egbib}
}

\end{document}